\documentclass{article}
\usepackage{spconf,amsmath,graphicx}
\usepackage{mdframed}
\usepackage{amssymb}
\usepackage{amsthm}
\usepackage{enumerate}
\usepackage{blindtext}
\usepackage{multirow}
\usepackage{multicol}
\usepackage{graphicx}
\usepackage{placeins}
\usepackage{caption}
\usepackage{algorithm}
\usepackage{algpseudocode}
\usepackage{mathrsfs}
\usepackage{tikz}
\usetikzlibrary{shapes,fit,positioning,arrows}
\usepackage[utf8x]{inputenc}
\usepackage{scalefnt}
\usepackage{epstopdf}

\usetikzlibrary{bayesnet}
\usetikzlibrary{shapes,arrows,backgrounds,positioning}

\DeclareMathOperator*{\argmin}{arg\,min}

\usepackage{subcaption}
\usepackage{multirow}
\usepackage{bbm}

\title{Re-Weighted Learning for Sparsifying Deep Neural Networks}
\name{Igor Fedorov \sthanks{Igor Fedorov was partially supported by the San Diego Chapter of the ARCS Foundation, Inc.}, Bhaskar D. Rao}
\address{Department of Electrical and Computer Engineering\\ University of California, San-Diego}

\begin{document}
    
\maketitle
\begin{abstract}
This paper addresses the topic of sparsifying deep neural networks (DNN's). While DNN's are powerful models that achieve state-of-the-art performance on a large number of tasks, the large number of model parameters poses serious storage and computational challenges. To combat these difficulties, a growing line of work focuses on pruning network weights without sacrificing performance. We propose a general affine scaling transformation (AST) algorithm to sparsify DNN's. Our approach follows in the footsteps of popular sparse recovery techniques, which have yet to be explored in the context of DNN's. We describe a principled framework for transforming densely connected DNN's into sparsely connected ones without sacrificing network performance. Unlike existing methods, our approach is able to learn sparse connections at each layer simultaneously, and achieves comparable pruning results on the architecture tested.

\end{abstract}

\section{Introduction}
Deep neural networks (DNN's) have become popular in a large number of fields due to their flexibility, simple learning procedure, and performance \cite{krizhevsky2012imagenet}. At a high level, DNN's learn a mapping from a set of inputs to a set of desired outputs. More formally, let $D = \lbrace x_i,y_i \rbrace_{i=1}^N$ be a dataset consisting of input and target output pairs. The DNN learning problem can be stated as
\begin{align}\label{eq:baseline}
\argmin_{\theta} f\left(\theta, D\right)
\end{align}
where $\theta = \lbrace W_k,b_k \rbrace_{k=1}^K$ is the set of weights and biases, respectively, which parametrize each of the $K$ network layers and $f(\cdot,\cdot)$ is an application dependent objective function. In the following, we will omit the dependence of $f(\cdot,\cdot)$ on $D$ for brevity. Due to space limitations, we omit further background and details on DNN's and refer the reader to \cite{goodfellow2016deep}.

As the number of parameters in the network grows, the complexity of the learned mapping grows with it. In fact, it has been shown that a DNN with a single hidden layer and finite number of neurons can approximate any measurable function arbitrarily well \cite{hornik1989multilayer}\footnote{For the result to hold, the non-linearity must be a squashing function \cite{hornik1989multilayer}.}. From a practical point of view, performing inference with a large DNN presents various challenges, including excessive power consumption and memory requirements \cite{han2015learning}. As such, a growing trend in the DNN research community has been to try to prune trained models, i.e. throw away some network parameters without harming performance. Work on DNN pruning goes back at least several decades, with early papers focusing on identifying network weights which have small influence on the objective function as measured by the Hessian of $f(\cdot)$ (or its approximation) \cite{lecun1989optimal,hassibi1993second}. A recent Hessian-based technique extends \cite{lecun1989optimal} by ensuring that the difference in network output at each layer of the original and pruned models is bounded \cite{dong2017learning}.

Other works have shown that the magnitude of a network weight can be a viable measure of its importance. The general paradigm is to undertake an iterative search where, given an estimate of the network parameters at iteration $t-1$, $\theta^{t-1}$, to alternate between
\begin{align}
\theta^{(t-0.5)} &= \argmin_{\theta: f(\theta) = f(\theta^{(t-1)})} \Vert \theta \Vert_0 \tag{Pruning}\label{eq:prune}\\
\theta^t &= \argmin_{\theta: \Vert \theta \Vert_0 = \Vert \theta^{(t-0.5)} \Vert_0} f(\theta) \tag{Retraining}\label{eq:retrain}
\end{align}
where $\Vert \theta \Vert_0$ denotes the number of non-zero elements in $\lbrace W_k \rbrace_{k=1}^K$. For instance, the Learning both Weights and Connections (LWC) algorithm performs \eqref{eq:prune} by setting small weights to $0$ \cite{han2015learning}. The issue with LWC is that if a parameter is mistakenly pruned in the \eqref{eq:prune} step, that weight will never be spliced in future iterations\footnote{Splicing refers to re-introducing a pruned parameter \cite{guo2016dynamic}.}. To remedy this shortcoming, the Dynamic Network Surgery (DNS) algorithm proposes to replace $\theta$ in \eqref{eq:prune}-\eqref{eq:retrain} with $q \odot \omega^{t-1}$, where $\odot$ denotes element-wise multiplication, the elements of $q$ represent the value of the corresponding elements of $\theta$, and the elements of $\omega^{t-1}$ denote whether the corresponding element of $\theta$ should be pruned at iteration $t$ or not \cite{guo2016dynamic}\footnote{In this work, as in \cite{guo2016dynamic,han2015learning}, we are interested in pruning the network weights only, i.e. not the biases.}. More specifically, 
\begin{align}\label{eq:T}
\omega_j^{t} &= \begin{cases} 
      0 & \vert q_j^{t-1} \vert < a \\
      1 & \vert q_j^{t-1} \vert \geq b \\
      \omega_j^{t-1} & \text{else}
   \end{cases}
\end{align}
and $q_j$ refers to the $j$'th element of $q$\footnote{In \cite{guo2016dynamic}, each layer $k$ has its own ($a_k,b_k$), but we have omitted this detail in \eqref{eq:T} for brevity.}. In the context of DNS, the \eqref{eq:prune} stage consists of evaluating \eqref{eq:T} and the \eqref{eq:retrain} stage is replaced by
\begin{align}\label{eq:guo}
q^t = \argmin_{q} f \left( q \odot \omega^{t-1} \right).
\end{align}
The benefit of \eqref{eq:guo} over \eqref{eq:retrain} is that even if a given weight has been pruned, i.e. the corresponding element of $\omega^{t-1}$ is $0$, that weight will still be updated while solving \eqref{eq:guo} and may eventually exceed the threshold $b$ and be spliced. 

\subsection{Contribution}
While the subject of sparsity has only recently gained traction in the DNN community, a considerable amount of literature dedicated to sparse solutions of linear systems already exists in the signal processing community. The purpose of this paper is to begin to bridge the gap between the two fields and show that a popular class of sparse signal recovery (SSR) techniques can be transferred to the task of DNN pruning. In the following, we propose a general affine scaling transformation (AST) algorithm for sparsifying DNN's. Unlike LWC and DNS, which perform pruning layer by layer, our approach learns sparse connections at all layers simultaneously. In some sense, this makes the proposed approach less greedy and allows it greater flexibility in exploring the search space. We will show that this framework is general, gives rise to many effective approaches, and is related to the state-of-the-art DNS algorithm.

\section{Proposed Framework}
\label{sec:proposed}
Consider the regularized DNN learning problem
\begin{align}\label{eq:regularized}
\argmin_{\theta} f(\theta) + \lambda \sum_{j=1}^{J} g(\theta_j)
\end{align}
where $g(\cdot)$ is a sparsity promoting regularizer and $J$ denotes the number of elements of $\theta$. While \eqref{eq:regularized} provides the benefit of learning sparse $\theta$, the trade-off is that the solution of \eqref{eq:regularized} may not necessarily be a solution to \eqref{eq:baseline}. Suppose, further, that $g(\cdot)$ is a concave function. It can be shown that the objective in \eqref{eq:regularized} is non-increasing under the update rule
\begin{align}\label{eq:re-weighted}
\argmin_\theta f(\theta) + \lambda \sum_{j=1}^{J} \theta_j \odot \psi_j^{t-1}
\end{align}
where $\psi_j^{t-1} = \triangledown g(\theta_j^{t-1})$. Methods like \eqref{eq:re-weighted} are collectively known as majorization-minimization (MM) algorithms \cite{hunter2004tutorial}. Let $q = \theta \odot \psi_j^{t-1}$ and $\lambda = 0$, then \eqref{eq:re-weighted} becomes
\begin{align}\label{eq:AST}
\argmin_q f\left(q \odot \omega^{t-1} \right)
\end{align}
where $\omega_j = \left(\psi_j\right)^{-1}$. Let $q^t$ by the solution of \eqref{eq:AST}. The proposed approach proceeds in an iterative fashion, where each iteration consists of finding $q^t$ and computing 
\begin{align}\label{eq:theta}
\theta^t = q^t \odot \omega^{t-1}.
\end{align}
Unlike LWC and DNS, our learning procedure is global, i.e. all of the network weights are updated at each iteration. 

\subsection{Why Affine Scaling?}
The method in \eqref{eq:AST} is known as an AST algorithm. While AST algorithms have been studied in great depth in the signal processing and optimization communities \cite{gorodnitsky1997sparse,rao1999affine,kreutz1997general,rao1996affine,nash1996linear}, their use has been limited in the DNN literature \cite{trafalis1996neural}. Such methods have a number of favorable properties, which can be applied to the task of sparsifying DNN's. One of the advantages of the AST is that an appropriately defined $\psi$, such as $\psi_j^{t-1} = \left(\theta_j^{t-1}\right)^{-1}$, allows for re-centering\footnote{By re-centering, we mean positioning the unknowns in the middle of the search space, such as $q_j = 1 \; \forall j$.} the optimization variables \cite{nash1996linear}. In the context of gradient based methods, centered variables allow for larger learning rates, especially when the optimization problem is constrained. In fact, early work on AST training of DNN's showed promising results in terms of decreased learning time \cite{trafalis1996neural}.

Another advantage of solving \eqref{eq:AST} is that the solution is guaranteed to  also be a solution of \eqref{eq:baseline}, which is not true for $\eqref{eq:regularized}$ with $\lambda > 0$. At the same time, the regularizer still plays a role in determining the solution space. In other words, if there are multiple solutions to \eqref{eq:baseline}, then iteratively solving \eqref{eq:AST} will tend to produce sparse choices of $\theta$. While we do not claim that the preceding argument for the sparsity of solutions is rigorous in the case of DNN's, it is well known in the context of SSR problems that AST methods converge to sparse solutions \cite{gorodnitsky1997sparse}. One important distinction between SSR and DNN training is that, in the case of SSR, $f(\theta)$ admits multiple minimizers and AST methods move from one minimizer to the next in search of sparse solutions, whereas the existence of multiple minimizers is not a given for the DNN objective.

\subsection{Special Cases}
\label{sec:special cases}
To illustrate how broad the proposed framework is, we proceed by showing the many forms which \eqref{eq:AST} can take for various choices of $g(\theta_j)$ used in the SSR literature. To the best of our knowledge, none of the following AST approaches have been used in the context of sparsifying DNN's.

Let $g(\theta_j) = \log \left(\vert \theta_j \vert + \tau\right)$, where $\tau > 0$ \cite{candes2008enhancing}. Then, \eqref{eq:AST} reduces to what is referred to as a re-weighted $\ell_1$ algorithm:
\begin{align}\label{eq:reweighted l1}
\argmin_q f\left(q \odot \left(\vert \theta^{t-1} \vert + \tau\right) \right)
\end{align}
where $\vert \cdot \vert$ refers to taking the absolute value of the input.

Suppose, instead, that $g(\theta_j) = h(\theta_j^2) = \log \left( \theta_j^2  + \tau\right)$ and consider repeating the MM procedure described in Section \ref{sec:proposed} for $h(\cdot)$ \cite{chartrand2008iteratively}. Then, \eqref{eq:AST} becomes what is referred to as a re-weighted $\ell_2$ algorithm:
\begin{align}\label{eq:reweighted l2}
\argmin_q f\left(q \odot \sqrt{\left(\theta^{t-1}\right)^2 + \tau} \right).
\end{align}

Another variant of \eqref{eq:AST} comes from the FOCUSS algorithm \cite{gorodnitsky1997sparse}, which uses $g(\theta_j) = \vert \theta_j\vert^p, 0 \leq p \leq 2$. In other words, FOCUSS considers $\ell_p$ norm regularization, which includes the $\ell_0$ pseudo-norm. Repeating the MM procedure for $g(\theta_j) = h(\theta_j^2) = \left(\vert \theta_j \vert^2\right)^{p/2}$, \eqref{eq:AST} becomes
\begin{align}\label{eq:focuss}
\argmin_q f\left(q \odot \left(\vert \theta^{t-1} \vert^{2-p} + \tau\right) \right)
\end{align}
where $\tau > 0$ is added for stability purposes \cite{gorodnitsky1997sparse}.

\subsection{Implementation Details}
\label{sec:implementation}
In practice, several considerations must be taken into account in the implementation of the proposed approach in \eqref{eq:AST}. Ideally, one would use \eqref{eq:AST} to find successively sparser estimates of $\theta$ while retaining the same network performance. We employ the stochastic gradient descent (SGD) algorithm and Theano software to find a stationary point of \eqref{eq:AST} at each re-weighting iteration \cite{2016arXiv160502688short}. In order to prevent instabilities in the propagation of gradients through the network, it is important that each re-weighting iteration $t$ is initialized such that the network is not taken too far from its state at $t-1$. For instance, one could initialize $q$ in \eqref{eq:AST} to $\theta^{t-1} \odot \psi_{j}^{t-1}$, but this would not allow the learning procedure to move to a new, sparser solution because the initializer would already be a stationary point of $f(\cdot)$ by definition. We propose two alternatives. The first option is to initialize $q$ using
\begin{align}\label{eq:initializer}
\left(\theta^{t-1} \odot  \psi_{j}^{t-1} \right) + v, v_j \sim \mathsf{N}(0,\sigma^2)
\end{align}
where $\sigma$ controls how far \eqref{eq:initializer} is from the previous state of the network. Setting $\sigma$ too large can result in instabilities, whereas setting $\sigma$ too small can result in \eqref{eq:AST} converging to $\theta^{t-1}$. The second approach, which we refer to as the greedy method, initializes $q$ at re-weighting iteration $t$ to $q^{t-1}$.

The complete algorithm pseudo-code is summarized in Algorithm \ref{alg: complete}. To speed up convergence, it is possible to update $ \psi_{j}^{t-1}$ for a single network layer at each re-weighting iteration.  In this regime, the learning procedure remains global since all of the network weights are still updated at each iteration. 

As will be shown in Section \ref{eq:AST}, executing Algorithm \ref{alg: complete} leads to a network whose weights are heavily concentrated around $0$, but not necessarily strictly equal to $0$. The task then becomes to select which weights to prune. We prune the weights at each layer by thresholding, re-train the entire network, and repeat the procedure for the rest of the layers (i.e. the LWC algorithm applied to the output of Algorithm \ref{alg: complete}). In this case, pruning based on magnitude is justified because the regularizer in \eqref{eq:regularized} pushes weights which do contribute to the minimization of $f(\cdot)$ toward $0$. Moreover, splicing operations like the ones employed by DNS are unnecessary. 
 
\begin{algorithm}
\begin{algorithmic}[1]
\Require $\theta^0$
\State $t \leftarrow 1$
\While{not converged}
\State Compute $ \psi_{j}^{t-1}$
\State Solve \eqref{eq:AST} to obtain $q^t$
\State Update $\theta^t$ using \eqref{eq:theta}
\State $t \leftarrow t + 1$
\EndWhile
\State \Return $\theta^{t+1}$
\end{algorithmic}
\caption{Proposed algorithm}
\label{alg: complete}
\end{algorithm}

\begin{figure*}
\centering
\begin{subfigure}[b]{0.23\textwidth}
\centering \includegraphics[width=\columnwidth]{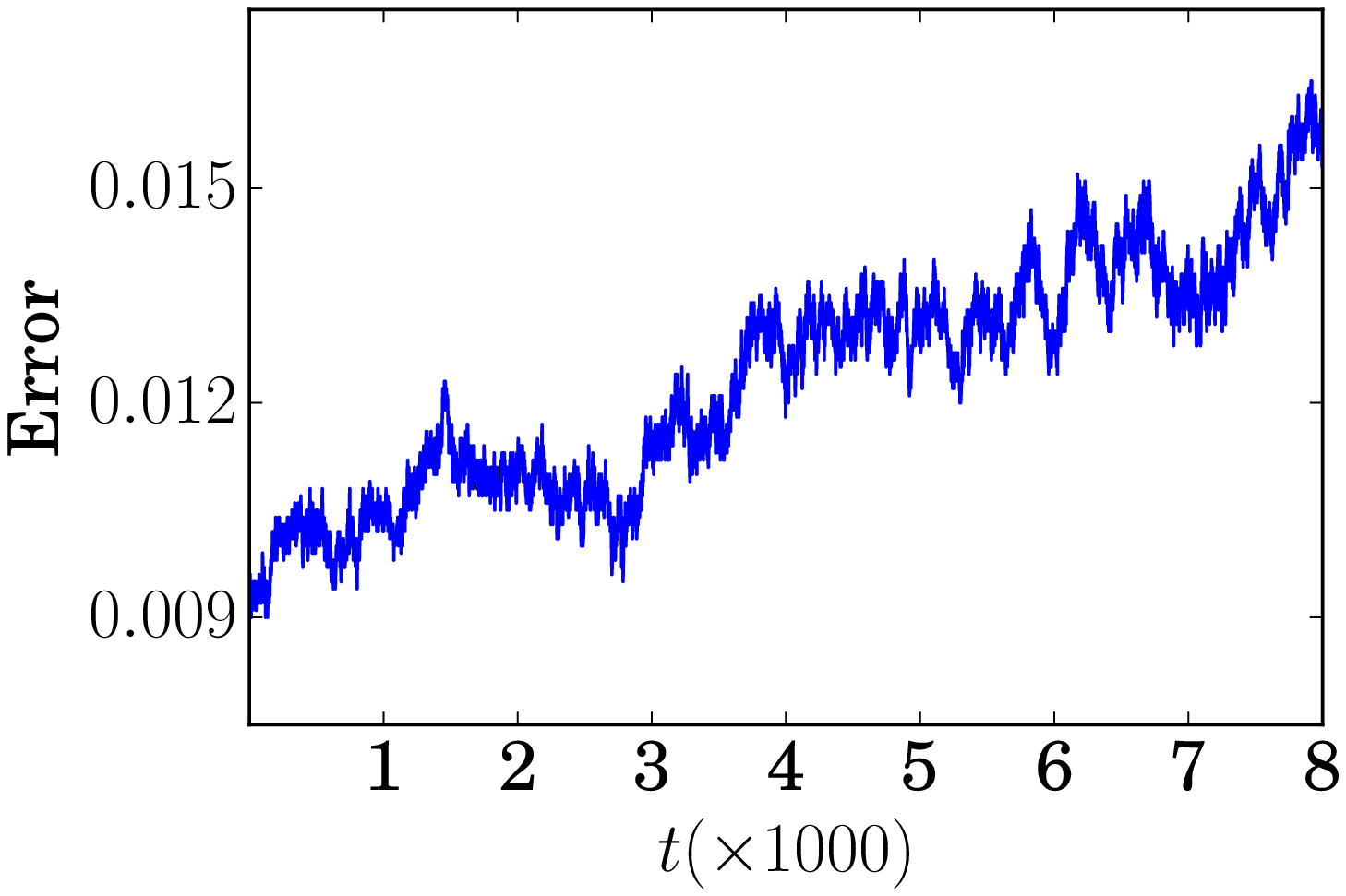}
\caption{\label{fig:gradual error}}
\end{subfigure}
~
\begin{subfigure}[b]{0.23\textwidth}
\centering \includegraphics[width=\columnwidth]{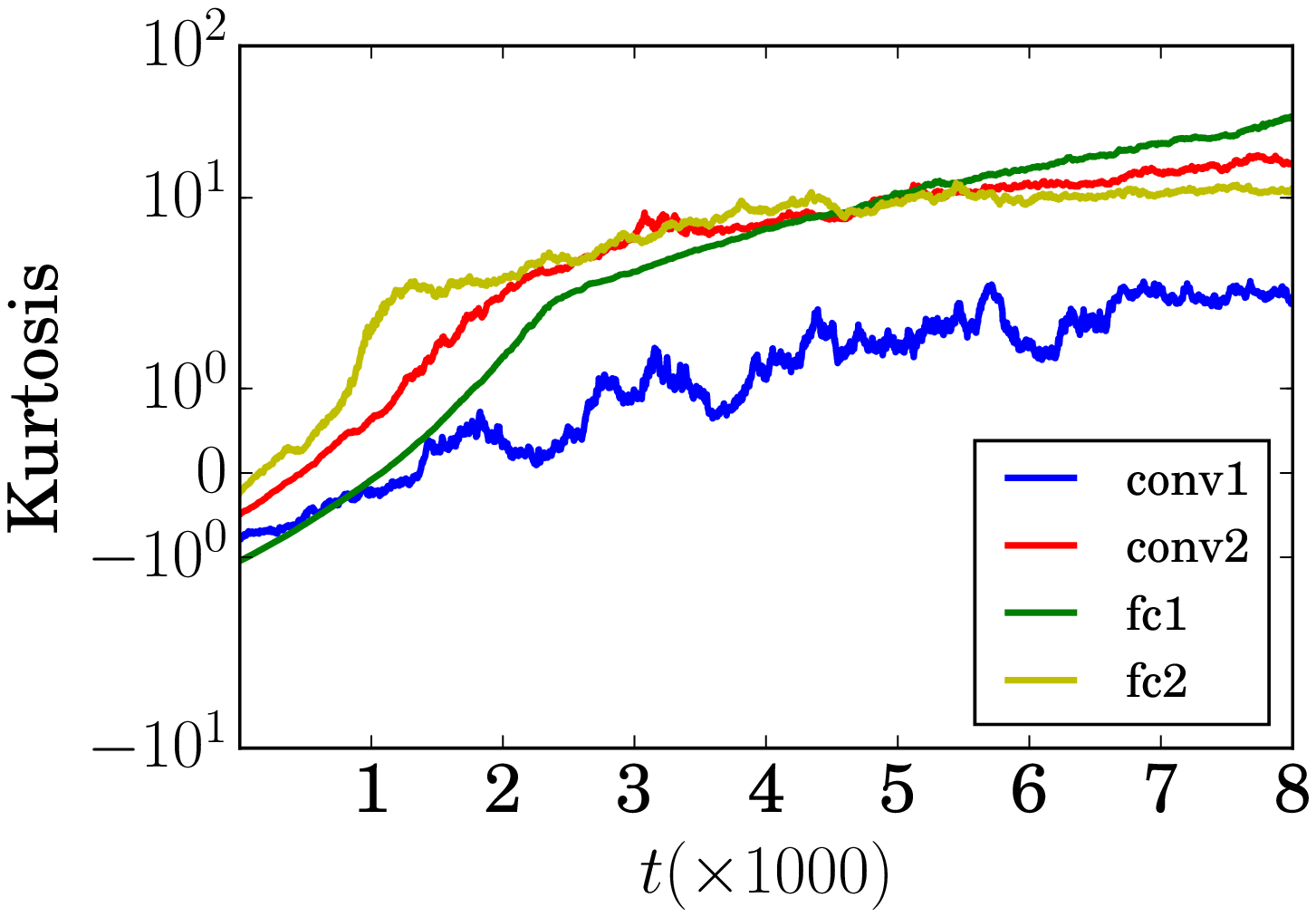}
\caption{\label{fig:gradual kurtosis}}
\end{subfigure}
~
\begin{subfigure}[b]{0.23\textwidth}
\centering \includegraphics[width=\columnwidth]{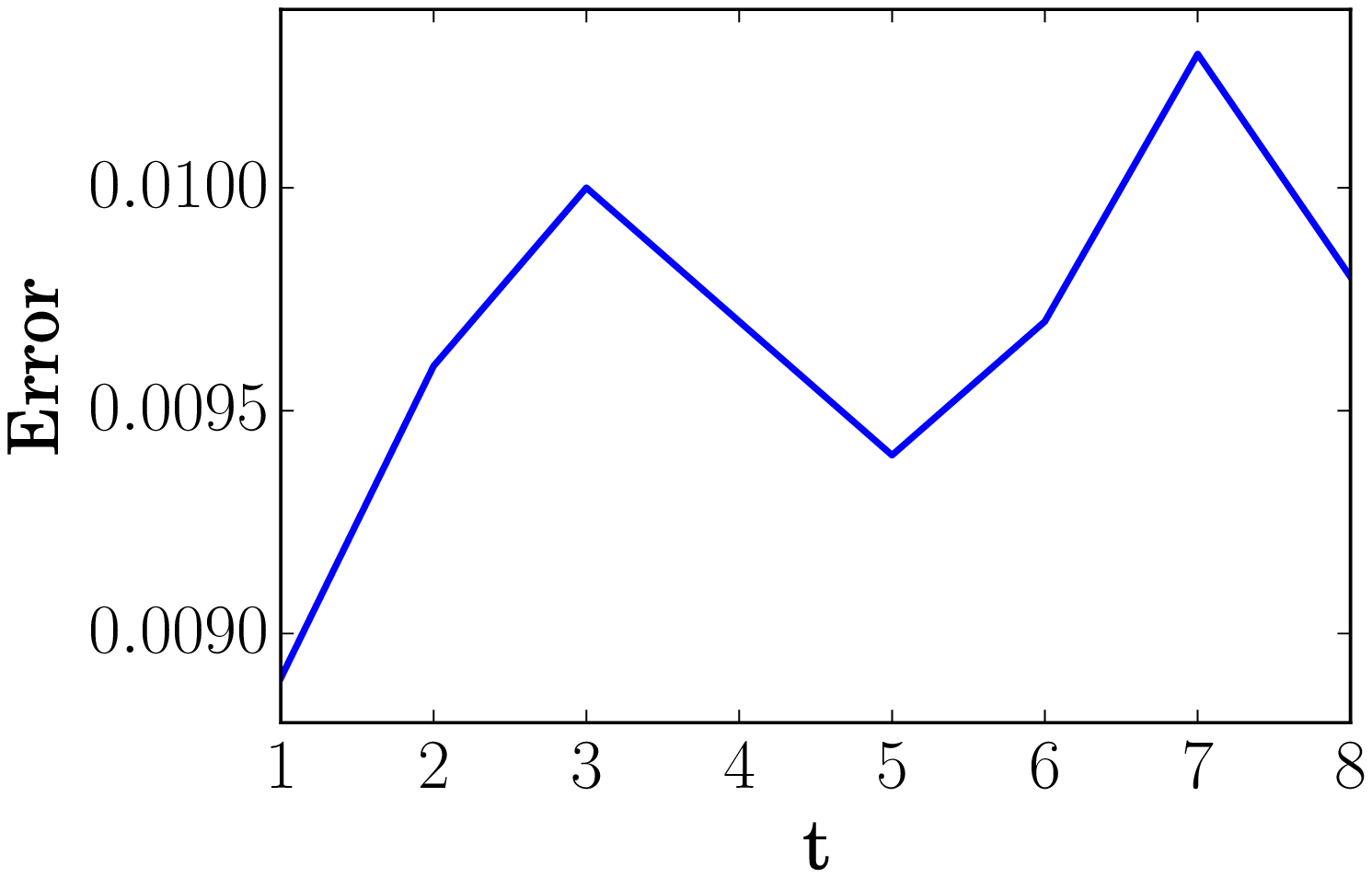}
\caption{\label{fig:greedy error}}
\end{subfigure}
~
\begin{subfigure}[b]{0.23\textwidth}
\centering \includegraphics[width=\columnwidth]{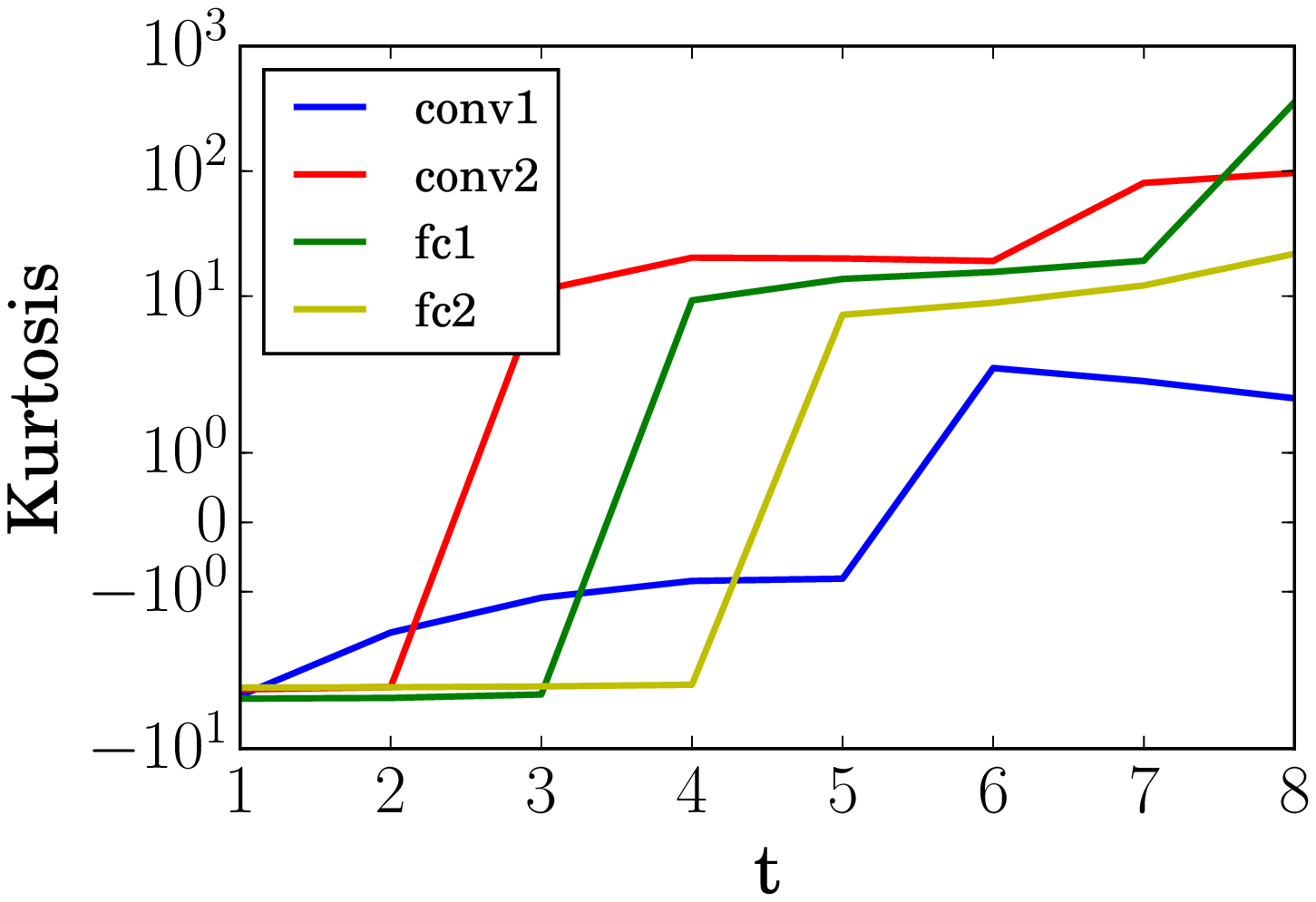}
\caption{\label{fig:greedy kurtosis}}
\end{subfigure}
\caption{Visualization of proposed learning procedure using the AST in \eqref{eq:reweighted l1}. \ref{fig:gradual error}-\ref{fig:gradual kurtosis} show the evolution of validation set error and kurtosis as a function of $t$ for the $Q$ initializer in \eqref{eq:initializer}. \ref{fig:greedy error}-\ref{fig:greedy kurtosis} show the evolution of validation set error and kurtosis as a function of $t$ for the greedy $Q$ initializer.}
\label{fig:greedy}
\end{figure*}

\begin{table*}
\centering
\begin{tabular}{cc|ccc|cc}
& & \multicolumn{3}{ | c |}{Proposed methods} & & \\
& Reference &  Re-weighted $\ell_1$ & Re-weighted $\ell_2$ & FOCUSS ($p=0.5$) & DNS \cite{guo2016dynamic} & LWC \cite{han2015learning}\\
Conv1 & -- & 27.8 & 50.4 & 67.6 & 14.2& 66 \\ 
Conv2 &-- & 6 & 4.9 & 8.1 & 3.1 & 12 \\
FC1 &-- & 0.7 & 0.9 &1 &0.7 & 8 \\
FC2 &-- & 18.6 & 4.7 & 15 & 4.3 & 19 \\
Total & --& 1.28 & 1.26 & 1.7 & 0.9 & 8 \\
Test set error ($\%$) & 0.86 & 1.16  & 1.41 & 1.13 & 0.91 & 0.77
\end{tabular}
\caption{Pruning performance on LeNet-5 in terms of the $\%$ non-zeros and the test set error. }
\label{table:results}
\end{table*}

\subsection{Relation to Dynamic Network Surgery \cite{guo2016dynamic}}
Although the authors of \cite{guo2016dynamic} did not frame DNS as an AST approach, DNS can be seen as a special case of the proposed framework. Let $a = b$ in \eqref{eq:T} and 
\begin{align}\label{eq:g dns}
g(\theta_j) = u(\theta_j - a) \theta_j
\end{align}
where $u(\cdot)$ denotes the unit-step function. Then, it can be shown that \eqref{eq:AST} reduces to the DNS algorithm, with the exception that DNS computes $\omega$ using the scaled variable $q$ whereas the proposed framework uses $\theta$. Since $\omega$ is a binary variable, the only difference between the two approaches is when $\omega_j^{t-1} = 0$. In this case, $\omega^t$ must be $0$ for the proposed framework, implying that pruned connections stay pruned for the choice of $g(\theta_j)$ in \eqref{eq:g dns}. Notice that this discrepancy is a result of the choice of $g(\cdot)$ in \eqref{eq:g dns}. For the choices of $g(\cdot)$ in Section \ref{sec:special cases}, $\omega_j^t$ is guaranteed to be strictly greater than $0$.
 
\section{Results}
\label{sec:results}
This section presents experimental results for the proposed algorithms. We focus on classifying the MNIST dataset using LeNet-5, a convolutional neural network architecture consisting of two convolution layers and two fully connected layers, denoted conv1/conv2 and fc1/fc2, respectively \cite{lecun1998gradient}. We solve \eqref{eq:baseline} to obtain $\theta^0$, which has a total of $431\times 10^3$ parameters.

To begin, we show that the proposed framework allows for globally updating the network parameters, moving towards sparser solutions without sacrificing accuracy. We execute Algorithm \ref{alg: complete} using the re-weighted $\ell_1$ choice of $\omega$ in \eqref{eq:reweighted l1} and compare the performance of both initialization strategies. For the initializer in \eqref{eq:initializer}, we update $\omega^{t-1}$ for all network weights at each re-weighting iteration $t$ and run one SGD epoch to optimize \eqref{eq:AST} for each $t$. For the greedy initialization strategy, we update $\omega^{t-1}$ for only one layer at each $t$, running Algorithm \eqref{alg: complete} for $8$ iterations and executing $1000$ SGD epochs to optimize $\eqref{eq:AST}$ for each $t$\footnote{We update $\omega$ twice for a given layer during the learning process.}.  To measure the degree to which the distribution of the weights of each layer are sparse, we monitor the kurtosis
\begin{align}
E\left[ \left(\frac{W_k - \mu_k}{\sigma_k} \right)^4\right]
\end{align}
where $\mu_k$ and $\sigma_k$ denote the mean and standard deviation of the weights in $W_k$, respectively. Distributions with kurtosis greater than $3$ are called super-Gaussian, meaning that they have higher peaks at $0$ and heavier tails than the Gaussian. 
Fig. \ref{fig:greedy} shows the evolution of the kurtosis and validation set error as a function of $t$ for the two proposed $Q$ initialization strategies. For both strategies, the kurtosis increases with iteration number, while the validation set error remains nearly unchanged. This provides experimental evidence for the claim that Algorithm \ref{alg: complete} allows for moving towards sparser solutions of \eqref{eq:baseline}. In addition, Algorithm \ref{alg: complete} automatically embodies the intuition that the earlier convolution layers should be pruned less than the later fully connected layers. This can be seen in Fig. \ref{fig:greedy}, where the kurtosis of the convolutional layers is much smaller than that of the fully connected layers. Comparing Fig. \ref{fig:gradual error}-\ref{fig:gradual kurtosis} to Fig. \ref{fig:greedy error}-\ref{fig:greedy kurtosis}, the greedy initialization approach leads to much sparser solutions without sacrificing classification accuracy.

The pruning results are reported in Table \ref{table:results}. As described in Section \ref{sec:implementation}, we use thresholding to prune the network after running Algorithm \ref{alg: complete}. The proposed approaches are compared with DNS and LWC. We use the greedy initialization strategy and run Algorithm \ref{alg: complete} for $8$ iterations. For the re-weighted $\ell_2$ approach, we use the annealing strategy described in \cite{chartrand2008iteratively}, where $\tau$ is gradually decreased with increasing iteration number. The results show that the proposed methods are competitive with existing state-of-the-art approaches without requiring splicing operations, supporting the claim that our framework automatically pushes unimportant parameters toward $0$.

\section{Conclusion}
We have described a general AST approach for sparsifying DNN's. Our approach is founded in principles from the SSR literature and provides an effective method of increasing the sparsity of a given DNN without sacrificing performance. Our approach is competitive with state-of-the-art pruning approaches and has the distinct characteristic of learning sparse weights for the entire network simultaneously.

\bibliographystyle{IEEEbib}
\bibliography{../ManuscriptBib}
\end{document}